# Implementation of Continuous Bayesian Networks Using Sums of Weighted Gaussians

Eric Driver and Darryl Morrell
Telecommunications Research Center and
Department of Electrical Engineering
Arizona State University
Tempe, AZ 85287

## Abstract

Bayesian networks provide a method of representing conditional independence between random variables and computing the probability distributions associated with these random variables. In this paper, we extend Bayesian network structures to compute probability density functions for continuous random variables. We make this extension by approximating prior and conditional densities using sums of weighted Gaussian distributions and then finding the propagation rules for updating the densities in terms of these weights. We present a simple example that illustrates the Bayesian network for continuous variables; this example shows the effect of the network structure and approximation errors on the computation of densities for variables in the network.

## 1 INTRODUCTION

Bayesian networks provide a method of representing conditional independence relationships between random variables and of computing the probability distributions associated with these random variables. Bayesian networks were originally developed for discrete-valued random variables; there is an increasing interest in extending this approach to continuous-valued random variables. Previous work on networks of continuous-valued variables has required that the variables have Gaussian density functions and that the relationships between these variables be linear(Pearl 1988, Shachter 1989). Theoretical issues involving the representation of relationships between random variables using a network (directed or undirected graph) consisting of both continuous and discrete random variables, including tests for conditional independence, are addressed in (Lauritzen 1989, Lauritzen 1990); an approach to updating conditional probabilities in which a single Gaussian is used to approximate a mixture of Gaussian densities is given in (Spiegelhalter 1990). In this paper, we extend Bayesian networks to include random variables with arbitrary distributions. We make this extension by approximating prior and conditional densities using *sums of weighted Gaussian distributions*; to our knowledge, this is the first time this approximation technique has been used in Bayesian networks. Using this technique, we find propagation rules for updating the densities of the network variables; information propagates through the network in the form of messages consisting of weight, mean, and variance updates.

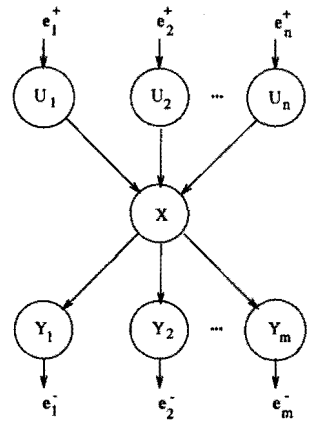

Figure 1: Fragment of a singly-connected tree.

## 2 PRELIMINARIES

We implement Bayesian networks for continuous variables by approximating both prior and conditional density functions by sums of weighted Gaussian densities. A density so approximated is represented in terms of a set of weights, a set of means, and a set of variances. Computation of these approximations is a hard problem which we discuss in Section 5.

Figure 1 shows a fragment of a Bayesian network. For our implementation, we assume that the network is singly connected (*i.e.* at most one path connects any two variables) and we assume that X is related to its parent nodes $U_1, U_2, \cdots, U_n$ by

$$X = g(U_1, U_2, \cdots, U_n) + w_x \qquad (1)$$

where $g$ is an arbitrary function and $w_x$ is a noise term assumed to be Gaussian with zero mean and uncorrelated with any other noise term or node in the network. In section 3.5 we consider the case when $g$ is linear.

From (1), we have:

$$f(x|u_1, \ldots, u_n) = N\left(x; \sigma_{w_x}^2, g(u_1, u_2, \cdots, u_n)\right) \qquad (2)$$

where $N(x; \sigma^2, \mu)$ is the Gaussian distribution defined by:

$$N(x; \sigma^2, \mu) \stackrel{\text{def}}{=} \frac{1}{\sqrt{2\pi\sigma^2}} \exp\left[\frac{-(x-\mu)^2}{2\sigma^2}\right] \qquad (3)$$

We approximate $f(x|u_1, \ldots, u_n)$ by a weighted sum of Gaussians as follows:

$$\begin{aligned} f(x|u_1, \ldots, u_n) &\approx \sum_{j=1}^{M} c_j N(x; \sigma_j^2, \mu_x^{(j)}) \cdot \\ &\quad \cdot N(u_1; \sigma_j^2, \mu_{u_1}^{(j)}) \cdots N(u_n; \sigma_j^2, \mu_{u_n}^{(j)}) \end{aligned} \qquad (4)$$

where $M$ is the number of Gaussians used to approximate $f(x|u_1, \ldots, u_n)$. Note that the approximation in (4) is capable of representing conditional density functions that do not embody the relationship between $X$ and $U_1, U_2, \ldots, U_n$ given in (1); thus, this approach can be used in more general settings than considered here. Again, we refer to Section 5 for a discussion of how the weights $\{c_j\}$, means $\left\{\mu_{u_k}^{(j)}\right\}$, and variances $\{\sigma_j^2\}$ can be determined to approximate a given conditional density function.

Using this approximation, we show that all belief functions and messages can be represented as sums of weighted Gaussians. Before proceeding, we state without proof the following identity which will be useful in the upcoming derivations:

$$\begin{aligned} N(x; \sigma_1^2, \mu_1) \cdot N(x; \sigma_2^2, \mu_2) &= \\ a \cdot N\left(x; \frac{\sigma_1^2 \sigma_2^2}{\sigma_1^2 + \sigma_2^2}, \frac{\mu_1 \sigma_2^2 + \mu_2 \sigma_1^2}{\sigma_1^2 + \sigma_2^2}\right) \end{aligned} \qquad (5)$$

where $a = N(\mu_1; \sigma_1^2 + \sigma_2^2, \mu_2)$ is a constant with respect to $x$.

## 3  THE PROPAGATION RULES

We first consider an arbitrary node $X$ shown in Figure 1. The parents of $X$ are $U_1, U_2, \ldots, U_n$ and the children of $X$ are $Y_1, Y_2, \ldots, Y_m$. We assume to begin that $X$ has not been instantiated. The case when $X$ has been instantiated and also the case when $X$ has no parents or no children is considered in Section 3.4.

### 3.1  COMPUTATION OF BEL(X)

Let $\mathbf{e}_X^+$ denote the evidence in the subnetwork above $X$ and let $\mathbf{e}_X^-$ denote the evidence in the subnetwork below $X$. Also, let $\mathbf{e}_i^+(i = 1, 2, \ldots, n)$ denote the evidence coming to $X$ via node $U_i$, and let $\mathbf{e}_j^-(j = 1, 2, \ldots, m)$ denote the evidence coming to $X$ via node $Y_j$. We assume $X$ has received messages $\pi_X(u_i)(i = 1, 2, \ldots, n)$ from its parents; each message is a sum of weighted Gaussians:

$$\pi_X(u_i) \stackrel{\text{def}}{=} f(u_i|\mathbf{e}_i^+) = \sum_{k_i=1}^{M_i} \alpha_{k_i}^i N\left(u_i; (\sigma_{\pi,k_i}^i)^2, \mu_{\pi,k_i}^i\right), \qquad (6)$$

where $M_i$ is the number of Gaussian densities in the representation for $\pi_X(u_i)$, $\{\alpha_{k_i}^i\}$ is a set of real valued weights, $\left\{\left(\sigma_{\pi,k_i}^i\right)^2\right\}$ is a set of real valued variances, and $\left\{\mu_{\pi,k_i}^i\right\}$ is a set of real valued means. Similarly, we assume $X$ has received messages $\lambda_{Y_j}(x)(j = 1, 2, \ldots, m)$ from its children:

$$\lambda_{Y_j}(x) \stackrel{\text{def}}{=} f(\mathbf{e}_j^-|x)$$
$$= \begin{cases} \sum_{l_j=1}^{P_j} \beta_{l_j}^j N\left(x; (\sigma_{\lambda,l_j}^j)^2, \mu_{\lambda,l_j}^j\right) & \text{if } \mathbf{e}_j^- \neq \emptyset \\ 1 & \text{if } \mathbf{e}_j^- = \emptyset \end{cases} \qquad (7)$$

where $P_j$ is the number of Gaussian densities in the representation for $\lambda_{Y_j}(x)$, $\left\{\beta_{l_j}^j\right\}$ is a set of real valued weights, $\left\{\left(\sigma_{\lambda,l_j}^j\right)^2\right\}$ is a set of real valued variances, and $\left\{\mu_{\lambda,l_j}^j\right\}$ is a set of real valued means.

The belief function for $X$ is computed as follows:

$$\begin{aligned} BEL(x) &\stackrel{\text{def}}{=} f(x|\mathbf{e}) \\ &= f(x|\mathbf{e}_X^+, \mathbf{e}_X^-) \\ &= \alpha f(x|\mathbf{e}_X^+) f(\mathbf{e}_X^-|x) \\ &= \alpha \pi(x) \lambda(x), \end{aligned} \qquad (8)$$

where $\alpha = \frac{1}{f(\mathbf{e}_X^-|\mathbf{e}_X^+)}$ is a normalization constant.

We compute $\pi(x)$ as follows:

$$\begin{aligned} \pi(x) &\stackrel{\text{def}}{=} f(x|\mathbf{e}_X^+) \\ &= \int_{u_1} \cdots \int_{u_n} f(x|\mathbf{e}_X^+, u_1, \ldots, u_n) \cdot \\ &\qquad \cdot f(u_1, \ldots, u_n|\mathbf{e}_X^+) du_1 \cdots du_n \\ &= \int_{u_1} \cdots \int_{u_n} f(x|u_1, \ldots, u_n) \cdot \\ &\qquad \cdot \prod_{i=1}^{n} f(u_i|\mathbf{e}_i^+) du_1 \cdots du_n \end{aligned}$$

$$= \int_{u_1} \cdots \int_{u_n} \sum_{j=1}^{M} c_j N(x; \sigma_j^2, \mu_x^{(j)}) N(u_1; \sigma_j^2, \mu_{u_1}^{(j)})$$

$$\cdots N(u_n; \sigma_j^2, \mu_{u_n}^{(j)}) \cdot \prod_{i=1}^{n} \sum_{k_i=1}^{M_i} \alpha_{k_i}^i \cdot$$

$$\cdot N\left(u_i; (\sigma_{\pi,k_i}^i)^2, \mu_{\pi,k_i}^i\right) du_1 \cdots du_n$$

$$= \sum_{j=1}^{M} c_j N\left(x; \sigma_j^2, \mu_x^{(j)}\right) \sum_{k_1=1}^{M_1} \cdots \sum_{k_n=1}^{M_n}$$

$$\prod_{i=1}^{n} \alpha_{k_i}^i \int_{u_i} N\left(u_i; \sigma_j^2, \mu_{u_i}^{(j)}\right) \cdot$$

$$\cdot N\left(u_i; (\sigma_{\pi,k_i}^i)^2, \mu_{\pi,k_i}^i\right) du_i$$

$$= \sum_{j=1}^{M} \gamma_j N\left(x; \sigma_j^2, \mu_x^{(j)}\right), \qquad (9)$$

where we have defined:

$$\gamma_j \stackrel{\text{def}}{=} c_j \sum_{k_1=1}^{M_1} \cdots \sum_{k_n=1}^{M_n} \prod_{i=1}^{n} \alpha_{k_i}^i \int_{u_i} N\left(u_i; \sigma_j^2, \mu_{u_i}^{(j)}\right)$$

$$\cdot N\left(u_i; (\sigma_{\pi,k_i}^i)^2, \mu_{\pi,k_i}^i\right) du_i$$

$$= c_j \sum_{k_1=1}^{M_1} \cdots \sum_{k_n=1}^{M_n} \prod_{i=1}^{n} \alpha_{k_i}^i \cdot$$

$$\cdot N\left(\mu_{u_i}^{(j)}; \sigma_j^2 + (\sigma_{\pi,k_i}^i)^2, \mu_{\pi,k_i}^i\right)$$

$$= c_j \prod_{i=1}^{n} \sum_{k_i=1}^{M_i} \alpha_{k_i}^i N\left(\mu_{u_i}^{(j)}; \sigma_j^2 + (\sigma_{\pi,k_i}^i)^2, \mu_{\pi,k_i}^i\right)$$

$$\qquad (10)$$

We now compute $\lambda(x)$, but first we make some definitions. Let $\mathcal{C}$ denote the set of child nodes of $X$, let $R = \{j \in \mathcal{C} | e_j^- \neq \emptyset\}$, and let $r$ be the number of elements in $R$. Next, we relabel the child nodes so that nodes $Y_1$ through $Y_r$ correspond to the nodes with $e_j^- \neq \emptyset$. If $r = 0$ then $\lambda(x) \stackrel{\text{def}}{=} 1$. If $r = 1$ then $\lambda(x) \stackrel{\text{def}}{=} \lambda_{Y_r}(x)$. For $r \geq 2$, we compute $\lambda(x)$ as follows:

$$\lambda(x) \stackrel{\text{def}}{=} f(e_X^- | x) = \prod_{j=1}^{r} f(e_j^- | x)$$

$$= \prod_{j=1}^{r} \sum_{l_j=1}^{P_j} \beta_{l_j}^j N\left(x; (\sigma_{\lambda,l_j}^j)^2, \mu_{\lambda,l_j}^j\right)$$

$$= \sum_{l_1=1}^{P_1} \cdots \sum_{l_r=1}^{P_r} \left(\prod_{j=1}^{r} \beta_{l_j}^j\right) \cdot$$

$$\cdot \prod_{j=1}^{r} N\left(x; (\sigma_{\lambda,l_j}^j)^2, \mu_{\lambda,l_j}^j\right) \qquad (11)$$

Using the fact that a product of Gaussians is proportional to a Gaussian we can see that (11) is indeed a weighted sum of Gaussians. We refer the reader to (Driver 1995) for a complete derivation. We write $\lambda(x)$ in the following form:

$$\lambda(x) = \sum_{j_0=1}^{M_0} \alpha_{j_0} N\left(x; (\sigma_{\lambda,j_0})^2, \mu_{j_0}\right) \qquad (12)$$

where $M_0 = \prod_{i=1}^{r} P_i$, $\{\alpha_{j_0}\}$ is a set of real valued weights, $\{(\sigma_{\lambda,j_0})^2\}$ is a set of real valued variances, and $\{\mu_{j_0}\}$ is a set of real valued means.

We can now compute $BEL(x)$ by forming the product of (12) and (9). $BEL(x)$ can then be written as a weighted sum of Gaussians by using (5):

$$BEL(x) = \alpha \sum_{j=1}^{M} \sum_{j_0=1}^{M_0} \eta_{j,j_0} \cdot$$

$$N\left(x; \frac{\sigma_j^2 (\sigma_{\lambda,j_0})^2}{\sigma_j^2 + (\sigma_{\lambda,j_0})^2}, \frac{\mu_{j_0} \sigma_j^2 + \mu_x^{(j)} (\sigma_{\lambda,j_0})^2}{\sigma_j^2 + (\sigma_{\lambda,j_0})^2}\right)$$

$$\qquad (13)$$

where

$$\eta_{j,j_0} = \gamma_j \alpha_{j_0} N\left(\mu_x^{(j)}; \sigma_j^2 + (\sigma_{\lambda,j_0})^2, \mu_{j_0}\right) \qquad (14)$$

and $\alpha$ is a normalization constant chosen so that $\int BEL(x) dx = 1$. Integrating (13) with respect to $x$, it follows that

$$\alpha = \left(\sum_{j=1}^{M} \sum_{j_0=1}^{M_0} \eta_{j,j_0}\right)^{-1} \qquad (15)$$

### 3.2 TOP DOWN PROPAGATION

The message $\pi_{Y_j}(x)$ that node $X$ sends to its $j$th child ($j = 1, 2, \ldots, m$) is formed as follows:

$$\pi_{Y_j}(x) \stackrel{\text{def}}{=} f(x | e - e_j^-)$$
$$= BEL(x | e_j^- = \emptyset)$$
$$= \alpha \pi(x) \lambda(x)|_{\lambda_{Y_j}(x)=1} \qquad (16)$$

So $\pi_{Y_j}(x)$ can be computed by the method of the last section with the assumption that $\lambda_{Y_j}(x) = 1$.

### 3.3 BOTTOM UP PROPAGATION

The message $\lambda_X(u_i)$ that node $X$ sends to its $i$th parent ($i = 1, 2, \ldots, n$) is formed as follows:

$$\lambda_X(u_i) \stackrel{\text{def}}{=} f(e - e_i^+ | u_i)$$

$$= \int_{u_1} \cdots \int_{u_{i-1}} \int_{u_{i+1}} \cdots \int_{u_n} \int_x$$

$$f(e - e_i^+ | u_1, \ldots, u_n, x)$$

$$\cdot f(u_1, \ldots, u_{i-1}, u_{i+1}, \ldots, u_n, x | u_i)$$

$$\cdot dx du_1 \cdots du_{i-1} du_{i+1} \cdots du_n$$

$$\qquad (17)$$

Consider the first distribution in the integrand:

$$f(\mathbf{e} - \mathbf{e}_i^+ | u_1, \ldots, u_n, x)$$
$$= f(\mathbf{e}_1^-, \ldots, \mathbf{e}_m^-, \mathbf{e}_1^+, \ldots, \mathbf{e}_{i-1}^+, \mathbf{e}_{i+1}^+, \ldots, \mathbf{e}_n^+ | u_1, \ldots, u_n, x)$$
$$= f(\mathbf{e}_1^-, \ldots, \mathbf{e}_m^- | x) \cdot$$
$$\quad f(\mathbf{e}_1^+, \ldots, \mathbf{e}_{i-1}^+, \mathbf{e}_{i+1}^+, \ldots, \mathbf{e}_n^+ | u_1, \ldots, u_n, x)$$
$$= \lambda(x) \prod_{\substack{k=1 \\ k \neq i}}^n f(\mathbf{e}_k^+ | u_k)$$
$$= \lambda(x) \prod_{\substack{k=1 \\ k \neq i}}^n \frac{f(u_k | \mathbf{e}_k^+) f(\mathbf{e}_k^+)}{f(u_k)}$$
$$= C \lambda(x) \prod_{\substack{k=1 \\ k \neq i}}^n \pi_X(u_k) \prod_{\substack{k=1 \\ k \neq i}}^n \frac{1}{f(u_k)} \qquad (18)$$

Where $C = \prod_{\substack{k=1 \\ k \neq i}}^n f(\mathbf{e}_k^+)$ is constant for a given set of evidence. Next, consider the second distribution in the integrand:

$$f(u_1, \ldots, u_{i-1}, u_{i+1}, \ldots, u_n, x | u_i)$$
$$= f(x|u_1, \ldots, u_n) f(u_1, \ldots, u_{i-1}, u_{i+1}, \ldots, u_n | u_i)$$
$$= f(x|u_1, \ldots, u_n) \prod_{\substack{k=1 \\ k \neq i}}^n f(u_k) \qquad (19)$$

Substituting (18) and (19) into (17), we have:

$$\lambda_X(u_i) = C \int_{u_1} \cdots \int_{u_{i-1}} \int_{u_{i+1}} \cdots \int_{u_n} \int_x \lambda(x)$$
$$\cdot \prod_{\substack{k=1 \\ k \neq i}}^n \pi_X(u_k) f(x|u_1, \ldots, u_n)$$
$$\cdot dx\, du_1 \cdots du_{i-1}\, du_{i+1} \cdots du_n \qquad (20)$$

The constant $C$ will get absorbed during the normalization of $BEL(x)$, so we can ignore it. $f(x|u_1, \ldots, u_n)$, $\lambda(x)$, and $\pi_X(u_k)$ are given by (4), (12), and (6) respectively. Substituting these into (20) we have:

$$\lambda_X(u_i) = \int_{u_1} \cdots \int_{u_{i-1}} \int_{u_{i+1}} \cdots \int_{u_n} \int_x \sum_{j_0=1}^{M_0} \alpha_{j_0}$$
$$\cdot N\left(x; (\sigma_{\lambda,j_0})^2, \mu_{j_0}\right) \prod_{\substack{k=1 \\ k \neq i}}^n \sum_{j_k=1}^{M_k} \alpha_{j_k}^k$$
$$\cdot N(u_k; (\sigma_{\pi,j_k}^k)^2, \mu_{\pi,j_k}^k) \sum_{j=1}^m c_j N(x; \sigma_j^2, \mu_x^{(j)})$$
$$\cdot N(u_1; \sigma_j^2, \mu_{u_1}^{(j)}) \cdots N(u_n; \sigma_j^2, \mu_{u_n}^{(j)})$$
$$\cdot dx\, du_1 \cdots du_{i-1}\, du_{i+1} \cdots du_n$$

$$= \sum_{j_0=1}^{M_0} \sum_{j_1=1}^{M_1} \cdots \sum_{j_{i-1}=1}^{M_{i-1}} \sum_{j_{i+1}=1}^{M_{i+1}} \cdots \sum_{j_n=1}^{M_n} \sum_{j=1}^m c_j \alpha_{j_0}$$
$$\cdot \left( \prod_{\substack{k=1 \\ k \neq i}}^n \alpha_{j_k}^k \right) N(u_i; \sigma_j^2, \mu_{u_i}^{(j)}) \int_{u_1} \cdots \int_{u_{i-1}}$$
$$\int_{u_{i+1}} \cdots \int_{u_n} \int_x N\left(x; (\sigma_{\lambda,j_0})^2, \mu_{j_0}\right)$$
$$\cdot N(x; \sigma_j^2, \mu_x^{(j)}) \prod_{\substack{k=1 \\ k \neq i}}^n N(u_k; (\sigma_{\pi,j_k}^k)^2, \mu_{\pi,j_k}^k)$$
$$\cdot N(u_k; \sigma_j^2, \mu_{u_k}^{(j)}) dx\, du_1 \cdots du_{i-1}\, du_{i+1} \cdots du_n$$

$$= \sum_{j=1}^m \psi_j N(u_i; \sigma_j^2, \mu_{u_i}^{(j)}) \qquad (21)$$

where we have defined:

$$\psi_j \stackrel{\text{def}}{=} c_j \sum_{j_0=1}^{M_0} \sum_{j_1=1}^{M_1} \cdots \sum_{j_{i-1}=1}^{M_{i-1}} \sum_{j_{i+1}=1}^{M_{i+1}} \cdots \sum_{j_n=1}^{M_n} \alpha_{j_0}$$
$$\cdot \int_x N\left(x; (\sigma_{\lambda,j_0})^2, \mu_{j_0}\right) N(x; \sigma_j^2, \mu_x^{(j)}) dx$$
$$\cdot \prod_{\substack{k=1 \\ k \neq i}}^n \alpha_{j_k}^k \int_{u_k} N(u_k; \sigma_j^2, \mu_{u_k}^{(j)})$$
$$\cdot N\left(u_k; (\sigma_{\pi,j_k}^k)^2, \mu_{\pi,j_k}^k\right) du_k$$
$$= c_j \sum_{j_0=1}^{M_0} \sum_{j_1=1}^{M_1} \cdots \sum_{j_{i-1}=1}^{M_{i-1}} \sum_{j_{i+1}=1}^{M_{i+1}} \cdots \sum_{j_n=1}^{M_n} \alpha_{j_0}$$
$$\cdot N\left(\mu_{j_0}; \sigma_j^2 + (\sigma_{\lambda,j_0})^2, \mu_x^{(j)}\right)$$
$$\cdot \left( \prod_{\substack{k=1 \\ k \neq i}}^n \alpha_{j_k}^k N\left(\mu_{u_k}^{(j)}; \sigma_j^2 + (\sigma_{\pi,j_k}^k)^2, \mu_{\pi,j_k}^k\right) \right)$$
$$= c_j \left( \sum_{j_0=1}^{M_0} \alpha_{j_0} N\left(\mu_{j_0}; \sigma_j^2 + (\sigma_{\lambda,j_0})^2, \mu_x^{(j)}\right) \right)$$
$$\cdot \prod_{\substack{k=1 \\ k \neq i}}^n \sum_{j_k=1}^{M_k} \alpha_{j_k}^k N\left(\mu_{u_k}^{(j)}; \sigma_j^2 + (\sigma_{\pi,j_k}^k)^2, \mu_{\pi,j_k}^k\right).$$
$$\qquad (22)$$

Note that if $\lambda(x)$ is constant then from (20) we have:

$$\lambda_X(u_i) = C \int_{u_1} \cdots \int_{u_{i-1}} \int_{u_{i+1}} \cdots \int_{u_n} \int_x$$
$$\cdot \prod_{\substack{k=1 \\ k \neq i}}^n \pi_X(u_k) f(x|u_1, \ldots, u_n)$$

$$\cdot dx du_1 \cdots du_{i-1} du_{i+1} \cdots du_n$$
$$= C \prod_{\substack{k=1 \\ k \neq i}}^{n} \int_{u_k} \pi_X(u_k) du_k$$
$$= \text{constant}$$

Hence, if $\lambda(x) = 1$ then $\lambda_X(u_i) = 1$ for each $i$. So just like with discrete variables, evidence gathered at a node does not affect any spousal nodes until their common child node obtains evidence.

### 3.4 BOUNDARY CONDITIONS

If $X$ is a root node (i.e. a node with no parents) that has not been instantiated, then we set $\pi(x)$ equal to the prior density function $f(x)$. This prior distribution is approximated by a sum of weighted Gaussians.

If $X$ is a leaf node (i.e. a node with no children) that has not been instantiated, then we set $\lambda(x) = 1$. This implies that $BEL(x) = \pi(x)$.

If $X$ is an evidence node, say $X = x_0$, then we set $\lambda(x) = \delta(x - x_0) = N(x; 0, x_0)$ regardless of the incoming $\lambda$-messages. This implies that $BEL(x) = N(x; 0, x_0)$ as we would expect. Furthermore, for each $j$, $\pi_{Y_j}(x) = N(x; 0, x_0)$ is the message that node $X$ sends to its children (each child gets the same message in this case). The messages that node $X$ sends to its parents are still weighted sums of Gaussians.

### 3.5 SPECIAL CASE: LINEARITY

In this section, we consider the special case when the function $g$ of Equation (1) has the form

$$g(U_1, U_2, \cdots, U_n) = b_1 U_1 + b_2 U_2 + \cdots + b_n U_n \quad (23)$$

where the $b_i$'s are real numbers. This relationship was suggested in (Pearl 1988). With this we have

$$f(x|u_1, \ldots, u_n) = N\left(x; \sigma_{wX}^2, \sum_{i=1}^{n} b_i u_i\right). \quad (24)$$

We note that only the computations of $\pi(x)$ and $\lambda_X(u_i)$ involve the distribution $f(x|u_1, \ldots, u_n)$, so all other computations are the same as before. Furthermore, we can obtain closed form solutions for $\pi(x)$ and $\lambda_X(u_i)$ without invoking the approximation given by (4). To do so, we use the following identity:

$$\int_{x_1} \cdots \int_{x_n} \left(\prod_{i=1}^{n} N(x_i; \sigma_i^2, \mu_i)\right) \cdot$$
$$\cdot N\left(\sum_{i=1}^{n} b_i x_i; \sigma^2, \mu\right) dx_1 \cdots dx_n$$
$$= N\left(\mu; \sigma^2 + \sum_{i=1}^{n} b_i^2 \sigma_i^2, \sum_{i=1}^{n} b_i \mu_i\right) \quad (25)$$

For completeness we state the final result but omit the proof; we refer the interested reader to (Driver 1995):

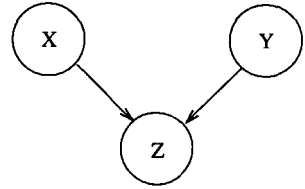

Figure 2: Network for the Example

$$\pi(x) = \sum_{k_1=1}^{M_1} \cdots \sum_{k_n=1}^{M_n} \left(\prod_{i=1}^{n} \alpha_{k_i}^i\right) \cdot$$
$$N\left(x; \sigma_{wX}^2 + \sum_{i=1}^{n} b_i^2 (\sigma_{\pi,k_i}^i)^2, \sum_{i=1}^{n} b_i \mu_{\pi,k_i}^i\right)$$
(26)

$$\lambda_X(u_i) = \sum_{j_0=1}^{M_0} \sum_{j_1=1}^{M_1} \cdots \sum_{j_{i-1}=1}^{M_{i-1}} \sum_{j_{i+1}=1}^{M_{i+1}} \cdots \sum_{j_n=1}^{M_n} \frac{\alpha_{j_0}}{|b_i|} \cdot$$
$$\left(\prod_{\substack{k=1 \\ k \neq i}}^{n} \alpha_{j_k}^k\right) N\left(u_i; \frac{1}{b_i^2}\left(\sigma_{wX}^2 + (\sigma_{\lambda,j_0})^2 + \sum_{\substack{k=1 \\ k \neq i}}^{n} b_k^2(\sigma_{\pi,j_k}^k)^2\right), \frac{\mu_{j_0}}{b_i} - \frac{1}{b_i}\sum_{\substack{k=1 \\ k \neq i}}^{n} b_k \mu_{\pi,j_k}^k\right)$$
(27)

Note that the means and variances of the Normal distributions in (26) and (27) have the same form as those in Pearl's result. The only real difference here is that our result is a sum of weighted Gaussians and Pearl's result is a single Gaussian.

## 4 EXAMPLE

In this section, we present a simple example that illustrates the characteristics of this approach to continuous Bayesian networks. The example uses the network of Figure 2. Nodes $X$ and $Y$ are independent and uniformly distributed on $[0, 1]$. Node $Z$ is related to $X$ and $Y$ by the following equation:

$$Z = X + Y + w_Z,$$

where $w_Z$ is a Gaussian random variable with zero mean and variance 0.01; $w_Z$ is independent of $X$ and $Y$.

Figure 3 shows a graph of our approximation to the prior distributions of nodes $X$ and $Y$; this approximation consists of 20 Gaussians with uniformly spaced means, equal variances, and equal weights. The absolute error in this approximation is 9%. Other approximations could be found that provide a better representation of the densities with fewer Gaussians; potential methods by which this can be done are discussed in Section 5.

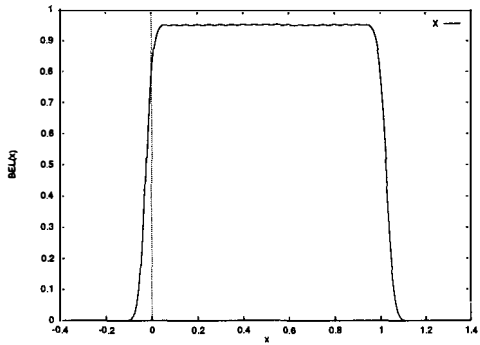

Figure 3: Prior pdfs of nodes X,Y

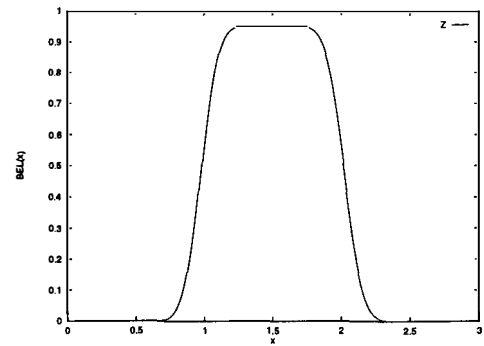

Figure 5: Node Z (Evidence: X=1.0)

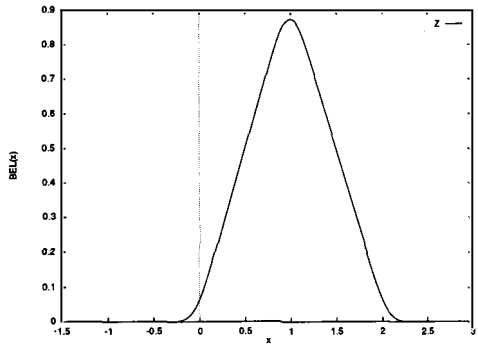

Figure 4: Node Z (No evidence)

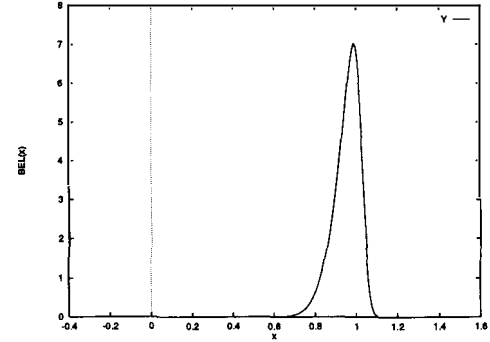

Figure 6: Nodes X,Y (Evidence: Z=2.0)

Performing the updating algorithm with no evidence in the network, we obtain the distribution for node $Z$. This is shown in Figure 4. From probability theory, we know that the distribution of a sum of two independent random varibles is the convolution of the individual distributions. So theoretically, the distribution of Z should be the triangular function on the interval $[0,2]$. Taking into account the effects of the noise $w_Z$ and the approximation error in the prior distributions, this is exactly what we have in Figure 4.

Next, we assume that node $X$ is instantiated at a value of 1.0. Performing the updating algorithm for this evidence, we obtain the belief function for node $Z$. This is shown in Figure 5. We can explain this as follows. Given $X = 1.0$, $Z$ is just the sum of the constant 1.0 and a uniform distribution (plus the noise term $w_Z$). Hence, $Z$ is a uniform distribution on $[1, 2]$. This is approximately what we have in Figure 5. Node $Y$ of course remains unchanged since it is independent of $X$.

Now we reinitialize the network and assume we have observed node $Z = 2.0$. Performing the updating algorithm for this evidence, we obtain the belief function for node $X$. This is shown in Figure 6. The only way that $Z$ can equal 2.0 is if both $X$ and $Y$ are 1.0, so the distributions for $X$ and $Y$ should be delta functions centered at 1.0. From Figure 6, we see that our results are a reasonable approximation to a delta function, especially when we consider the effects of the noise and the approximation error in the prior density functions.

## 5   GAUSSIAN SUM APPROXIMATIONS

In this section, we discuss briefly some of the issues involved in choosing sum-of-Gaussian approximations to probability density functions. We first discuss the existence of good approximations. We then mention several techniques that can be used to obtain these approximations.

Let $S$ be a compact subset of $\mathcal{R}^n$ and consider the set:

$$\mathcal{G}_S \;=\; \left\{ g \in C[S] \,\bigg|\, g(\mathbf{x}) = \sum_{i=1}^{m} c_i \exp\left[\frac{-1}{2\sigma_i^2}(\mathbf{x} - \boldsymbol{\mu}_i)^{\mathrm{T}} (\mathbf{x} - \boldsymbol{\mu}_i)\right] ; m \in \mathcal{N}; c_i, \sigma_i \in \mathcal{R}; \boldsymbol{\mu}_i \in \mathcal{R}^n \right\} \tag{28}$$

where $C[S]$ is the set of all continuous functions from $S$ to $\mathcal{R}$. By a simple application of the Stone-Weierstrass theorem, it can be shown (Girosi 1990) that $\mathcal{G}_S$ is dense in $C[S]$. That is, given any $f \in C[S]$ and any $\varepsilon > 0$, there exists $g \in \mathcal{G}_S$ such that $\int_S |g(\mathbf{x}) - f(\mathbf{x})| dV < \varepsilon$. (Note: here we have used the $L_1$ norm, in fact the result holds for any $L_p$ norm, $1 \leq p < \infty$.) In other words, any arbitrary function in $C[S]$ can be approximated arbitrarily well by a finite sum of weighted Gaussians.

In this paper, the functions we are approximating are probability density functions. Since density functions die off at infinity, as an approximation we may assume that they are defined on compact subsets of $\mathcal{R}^n$ (*i.e.* we set them to zero outside some compact subset of $\mathcal{R}^n$). Hence, by the above results it is reasonable to assume that any density function can be approximated arbitrarily well by a finite sum of Gaussians.

We now turn our attention to actually finding such approximations. One approach to obtaining approximations is the use of neural networks (Poggio 1990). Another method (Klopfenstein 1983) approximates an arbitrary function by uniformly spaced Gaussians; using Fourier transform techniques, error estimates can be obtained. Other methods include simmulated annealing and gradient descent algorithms. Our best results have been with the gradient descent algorithm, especially when approximating functions with a small number of variables.

### Acknowledgements

This work supported by the United States Army Research Office, grant number DAAH04-93-G-0218